\documentclass[runningheads]{llncs}
\usepackage{graphicx}
\usepackage{multirow}
\usepackage{booktabs}
\usepackage{array}
\usepackage{cite}
\usepackage{amsmath}
\usepackage{amssymb}
\usepackage[colorlinks, linkcolor=blue, anchorcolor=blue, citecolor=blue, urlcolor=black]{hyperref}
\usepackage{booktabs} 
\usepackage{ragged2e}
\usepackage{tabularx}
\usepackage[misc]{ifsym}
\authorrunning{First Author et al.}
\newcommand{\etal}{\textit{et al.}}

\begin{document}
\title{ProCo: Prototype-aware Contrastive Learning for Long-tailed Medical Image Classification}
%
%
\author{Zhixiong Yang\inst{1}\textsuperscript{*}  \and Junwen Pan\inst{1}\textsuperscript{*}  \and Yanzhan Yang\inst{1} \and Xiaozhou Shi\inst{1} \and Hong-Yu Zhou\inst{2} 
     \and Zhicheng Zhang\inst{1}\textsuperscript{$\dagger$} \and Cheng Bian\inst{1}\textsuperscript{$\dagger$}}
\institute{Xiaohe Healthcare, ByteDance, Guangzhou, China  \and
    Department of Computer Science, The University of Hong Kong, Pokfulam, Hong Kong
    \\
\email{\{biancheng, zc.zhang\}@bytedance.com} \\
    }

\authorrunning{Yang et al.}
\titlerunning{ProCo}
%
%
\maketitle              

\renewcommand{\thefootnote}{}
\footnotetext{\textsuperscript{$*$} Zhixiong Yang and Junwen Pan contributed equally to this work.}
\footnotetext{\textsuperscript{$\dagger$} Cheng Bian and Zhicheng Zhang are corresponding authors.}

\begin{abstract}
Medical image classification has been widely adopted in medical image analysis. However, due to the difficulty of collecting and labeling data in the medical area, medical image datasets are usually highly-imbalanced. To address this problem, previous works utilized class samples as prior for re-weighting or re-sampling but the feature representation is usually still not discriminative enough. In this paper, we adopt the contrastive learning to tackle the long-tailed medical imbalance problem. Specifically, we first propose the category prototype and adversarial proto-instance to generate representative contrastive pairs. Then, the prototype recalibration strategy is proposed to address the highly imbalanced data distribution. Finally, a unified proto-loss is designed to train our framework.
The overall framework, namely as \textbf{Pro}totype-aware \textbf{Co}ntrastive learning (\textbf{ProCo}), is unified as a single-stage pipeline in an end-to-end manner to alleviate the imbalanced problem in medical image classification, which is also a distinct progress than existing works as they follow the traditional two-stage pipeline. Extensive experiments on two highly-imbalanced medical image classification datasets demonstrate that our method outperforms the existing state-of-the-art methods by a large margin. Our source codes are available at \url{https://github.com/skyz215/ProCo}.

\end{abstract}
%

%
%
%
%
%
%
\section{Introduction}
Convolution neural network has been proved to be successful in many visual tasks \cite{he2016deep, ren2015faster, long2015fully, ronneberger2015u, isensee2018nnu, Ji_2021_CVPR}.
Although impressive breakthrough has been achieved, recent advances are still driven by the balanced dataset in the corresponding tasks. 
However, in real-world medical practice, the acquired dataset often exhibits long-tail distribution \cite{liu2019large}, where the head categories dominate most of the data, whereas tailed categories only have a handful of samples. 
Such skewed dataset is commonly originated from the 
difficulty of collecting rare diseases, or insufficient annotation from the proficient expertise. 
For this reason, the well-trained model is prone to make decision bias towards head classes due to the numerical superiority, weakening the model performance on tailed classes.

To address the long-tailed imbalance, follow-up studies have been conducted in recent years. 
Common solutions include class-rebalancing~\cite{wang2020devil, estabrooks2004multiple}, information supplement~\cite{wei2021crest, zang2021fasa}, and module modification~\cite{kang2019decoupling, zhou2020bbn, kang2020exploring}. Unfortunately, these works are heavily depended on manual designs or prior knowledge, causing low efficacy and poor generalization of their proposed models.
To this end, we propose \textbf{Pro}totype-aware \textbf{Co}ntrastive learning (\textbf{ProCo}) to address the long-tailed problem with decent performance and high efficiency.

Different from previous works, with the assistance of well-designed loss function Proto-loss, the main innovation of ProCo is that the proposed framework is a combination of the contrastive learning, category prototype, and proto-instance and can commendably tackle the long-tailed medical image classification.
%
%
%
Formally, our technical contributions can be summarized in four-fold:

\begin{itemize}
\item[$\bullet$] We propose a category prototype and adversarial proto-instance for feature modeling. Specifically, the category prototype can model the arbitrary category distribution adaptively.  Especially, adversarial proto-instance is generated from category prototype and representative instance to enhance the robustness of contrastive learning over all classes in the long-tailed setting.
\item[$\bullet$] We present a prototype recalibration strategy to ensure the updated frequency on category prototypes of tailed classes and eliminate the prototype bias, which is an imbalance updating process resulting in an incorrect distribution prediction.
\item[$\bullet$] To unify the contrastive learning together with prototype-based supervised learning, we propose a proto-loss, which can significantly boost the efficiency of our end-to-end framework ProCo.
\item[$\bullet$] Extensive experiments on two long-tailed medical classification datasets show that ProCo yields the best over state-of-the-arts by a large margin, demonstrating the effectiveness of the proposed method.
\end{itemize}

\section{Related Work}

Existing practices on the class-rebalancing aim to adjust the distribution of training samples to achieve balanced data over all categories, \textit{e.g.}, over-sampling~\cite{last2017oversampling} on tailed classes or under-sampling~\cite{koziarski2020radial} on head classes.
While these techniques have successfully improved the performance on tailed classes, the performance on head classes will be sacrificed~\cite{zhang2021deep}.
In this regards, information supplement methods are investigated by introducing prior expertise into the entire framework to overcome the performance degradation, \textit{e.g.}, transfer learning~\cite{weng2021unsupervised}, model pre-training~\cite{yang2020rethinking}, knowledge distillation~\cite{wang2020long}, and self-supervised learning~\cite{wei2021crest}.
Other studies explore module modification paradigms. 
For instance, Kang~\etal~\cite{kang2019decoupling} proposed a two-stage decoupling training method, in which the backbone and classifier were trained separately for the tailed class accommodation. 
Dong~\etal~\cite{dong2017class} proposed a metric learning approach for batch incremental hard sample mining of minority attribute classes from imbalanced large-scale training data.
However, one limitation of these methods is that the prior expertise relies on cumbersome manipulation, which is an obstacle to real-world application.
In this paper, we attempt to address the long-tail problem in a contrastive way, which can not only tackle above challenges but also achieve promising performance with competence of high efficiency.

\section{Methodology}

\begin{figure}[!t]
\centering
\includegraphics[width=\columnwidth]{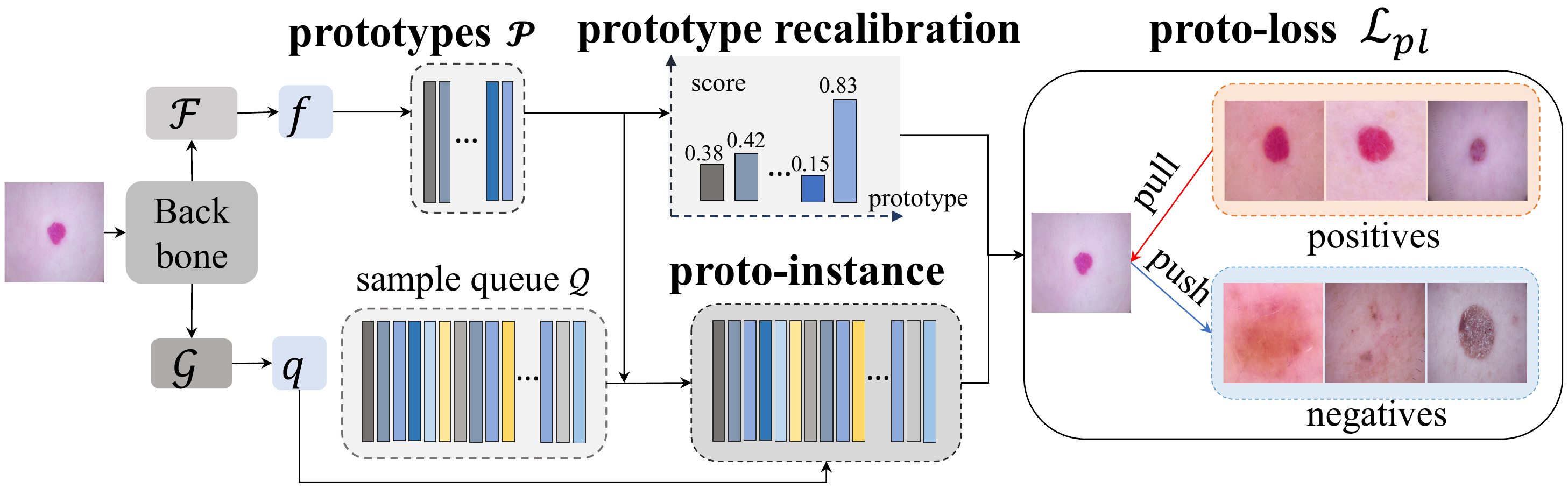}
\vspace{-0.2cm}
\caption{Our proposed Prototype-aware Contrastive Learning (ProCo) framework. $\mathcal{F}$ and $\mathcal{G}$ are two projectors, $f$ is the image feature, $q$ is the query feature. Prototypes $\mathcal{P}$ and sample queue $\mathcal{Q}$ are used to generate adversarial proto-instances. All negatives and positives enter into the proposed proto-loss $\mathcal{L}_{pl}$ for optimization. Besides, a prototype recalibration strategy is used for adjusting the weight of each prototype in proto-loss.}
\label{fig:framework}
\end{figure}

Fig. \ref{fig:framework} demonstrates the diagram of our proposed ProCo framework. It originates from MoCo~\cite{he2020momentum}, which consists of an online encoder and a momentum updated encoder. 
Here we omit the momentum encoder for simplicity.
As these two encoders share the same architecture, in the following text we only describe the structure of the online encoder. 

\subsubsection{Notation} The long-tailed classification training set with $N$ training samples and $C$ categories is denoted as $\mathcal{X} = \{(x_j, y_j)  \vert 1 \le j \le N \}$.
The network consists of a shared backbone and two projection heads notated as $\mathcal{G}$ and $\mathcal{F}$, respectively. 
The feature produced by projection head $\mathcal{F}$ is used for classification, while $\mathcal{G}$ for the contrastive learning.
The sample queue $\mathcal{Q}=\{q_j\in \mathbb{R}^{k} \vert 1\le j \le M\}$ with $M$ instances is maintained via the momentum fashion~\cite{he2020momentum}.

\subsection{Category Prototype and Adversarial Proto-instance}

\begin{figure}
\centering
\includegraphics[width=0.8\textwidth]{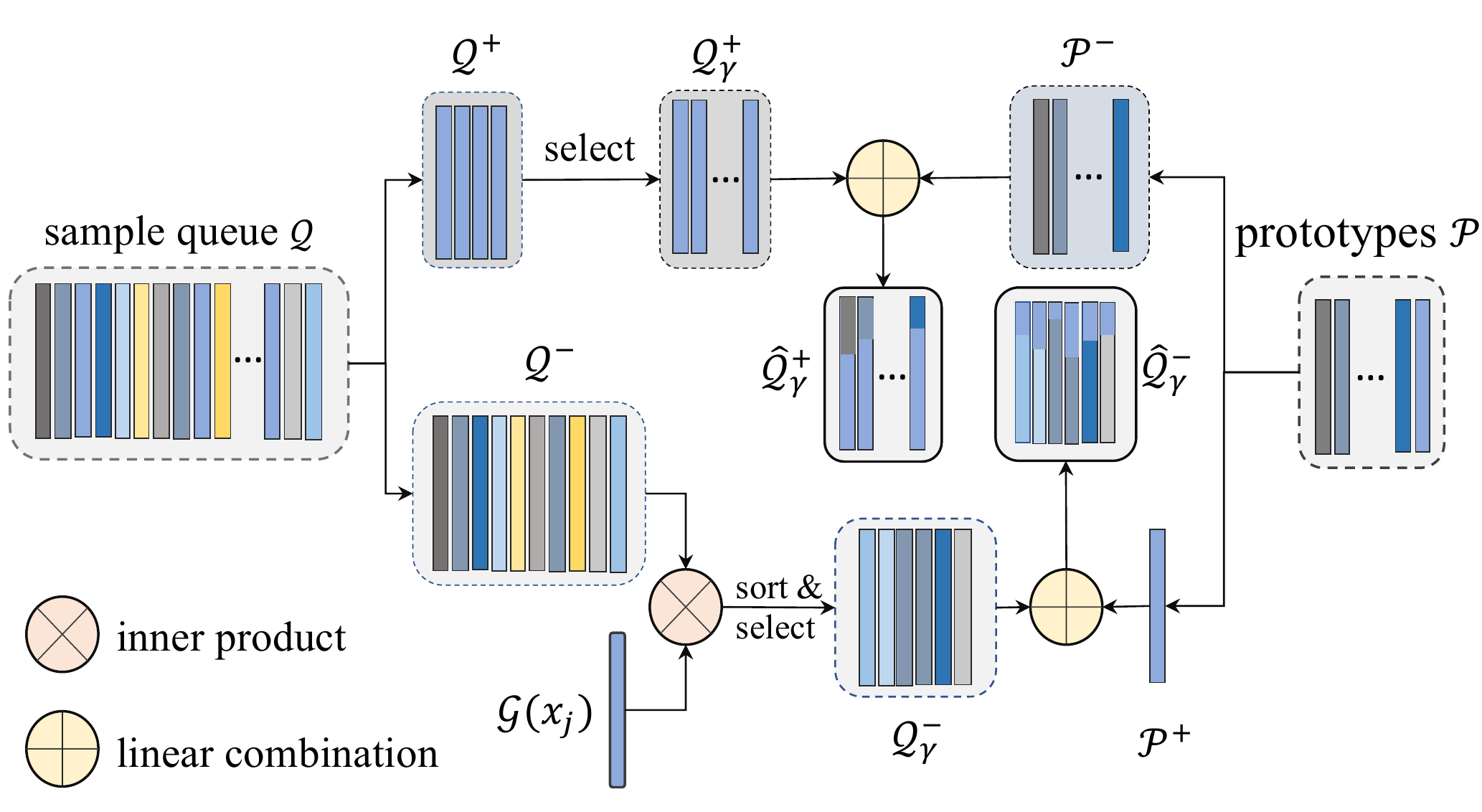}
\caption{Illustration of the proposed adversarial proto-instance. }
\vspace{-0.5cm}
\label{fig:mixup}
\end{figure}


Classic contrastive training pairs (\textit{i.e.}, positive and negative pairs) are used to learn the representation of instances. However, in the long-tailed dataset, the head classes dominate most of negative pairs via the conventional contrastive methods, causing the under-learning of tailed classes. Previous works \cite{cai2020all, robinson2020contrastive} reveal that not all negative pairs facilitate the contrastive learning. Therefore, the key to improve the performance of the long-tailed problem is to reduce the redundancy of contrastive negative pairs and mine recognizable positive pairs.
To this end, we propose a new concept named category prototype. The prototype is a set of learnable parameters $\mathcal{P}=\{p_c \in \mathbb{R}^{k} \vert 1 \le c \le C\}$ for predefined $C$ categories and is optimized by our proto-loss as described in Sec.~\ref{sec:loss}. 
Then, we generate the adversarial proto-instance from the category prototype and representative sample, \textit{i.e.}, confusing samples from the alternative sample queue, via a linear interpolation in the feature space, which will be utilized to form a training pair in ProCo. Theoretically, the adversarial proto-instance is designed as a special outlier, which can encourage ProCo to rectify the decision boundaries of the tailed categories during the contrastive learning.


%
Fig. \ref{fig:mixup} illustrates the diagram of the adversarial proto-instance. First, for each instance $(x_j, y_j)$, the queue $\mathcal{Q}$ is divided into two disjoint subsets $\mathcal{Q}^+ = \{ s_i | y_i=y_j \}$ and $\mathcal{Q}^- = \{ s_i | y_i \ne y_j \}$ that comprise positive and negative instances, respectively.
Similarly, $\mathcal{P}$ is also grouped into a singleton set $\mathcal{P}^+=\{p_c^+ | c=y_j \}$ with exactly one prototype $p_c^+$ and a negative set $\mathcal{P}^-=\mathcal{P} \setminus \mathcal{P}^+$, where $p_c^+$ is the positive prototype feature corresponding to the current instance.
Then, the adversarial positive proto-instances are derived from $\mathcal{P}^-$ and $\mathcal{Q}^+$ while the negative ones are from $\mathcal{P}^+$ and $\mathcal{Q}^-$.

To synthesize the adversarial negative proto-instances, we prioritize those negative instances that are likely to be confused with the current instance.
The distance between the current feature $\mathcal{G}(x_j)$ and negative feature $s^-_i \in \mathcal{Q}^-$ can be utilized as an indicator, and is represented as:
\vspace{-2mm}
\begin{equation}
d(x_j, s_i^-)= 1- \frac{\mathcal{G}(x_j)^\top \cdot s^-_i}{\Vert \mathcal{G}(x_j) \Vert_2 \Vert s^-_i\Vert_2 },
\end{equation}
where the superscript $\top$ represents the transpose operation. 
Then, we rank the negative instances in $\mathcal{Q}^-$ in ascending order according to their distance to $\mathcal{G}(x_j)$ and select the top $\gamma$ instances to compose the adversarial proto-instance set $\mathcal{Q}^-_{\gamma}$:
\vspace{-2mm}
\begin{equation}
    \mathcal{Q}^-_{\gamma} = \left\{s_i^- \vert s_i^- \in \mathcal{Q}^-, d(x_j, s_i^-)\le d(x_j, s_\gamma^-)\right\},
\end{equation}
where $s_\gamma$ is the $\gamma$-th element in the sorted $\mathcal{Q}^-$.
Finally, for more challenging negative instances, we randomly perturb each element in $\mathcal{Q}^-$ with the positive prototype $p_c^+$:
\begin{equation}
\hat{\mathcal{Q}}^{-}_\gamma = \left\{ \frac{(1-\epsilon_i) s_i + \epsilon_i p_c^+}{\Vert (1-\epsilon_i) s_i + \epsilon_i p_c^+ \Vert_2}  \ \Big\vert \ s_i \in \mathcal{Q}_\gamma^- \right\},
\label{eq:mixup}
\end{equation}
where $\epsilon_i \in (0, E)$ is a random interpolation coefficient for each sample and the upper bound $E$ is a hyperparameter with a small value.
We assume that prototype $p_c^+$ always contributes less than $s_i$ to the generated proto-instance, which guarantees that the negative semantic within $s_i$ can be held.

As for the adversarial positive proto-instances, the strategy is to select samples in $\mathcal{Q}^+$ that are misclassified and combine them with the incorrectly assigned prototype, where the interpolation manner is identical to that of Eq.~\ref{eq:mixup}.

\subsection{Prototype Recalibration}

Although the problem of constructing contrastive pairs has been addressed by the proposed category prototype and adversarial proto-instance, the learning of category prototype is still potentially affected by the class imbalance problem. We argue that the underlying reason is the prototype bias, where the generated proto-instance will incline to head classes, jeopardizing the performance on tailed classes. 
For this reason, this work proposes a prototype recalibration strategy, which estimates the representative level of each category prototype showing the importance of tailed class prototype features.

Specifically, inspired by the similarity between the projected features and prototype, we introduce a rectified sigmoid function to achieve recalibration. 
In formal,
the calibration factor for each category prototype $p_c\in \mathcal{P}$ is defined as:
\vspace{-2mm}
\begin{equation}
\omega _c \left(p_c, \{x_j\ |\ y_j=c\} \right) = \frac{1}{N_c}\sum_{1\le j\le N_c} \frac{1}{1+e^{-\mathcal{F}(x_j)^\top \cdot p_c} },
\end{equation}
where $\{x_j|y_j=c\}$ is a subset of samples in the associated category $c$, and $N_c$ is the total number of them. {\color{blue}{}}
To allow end-to-end calibration for batch-based training, we keep a running mean via the exponential moving average for the global calibration factor:
\vspace{-2mm}
\begin{equation}
    \bar\omega_c = \beta \cdot \bar\omega _c + (1 - \beta) \cdot \omega _c \left(p_c, \{x_j\ | \ y_j=c\} \right) ,
    \label{ema}
\end{equation}
where $\{x_j  |  y_j=c\}$ are samples in a batch with label $c$, and $\beta$ is a smoothing coefficient.

The calibration factor reflects the \textit{difficulty} of each category and the \textit{representativeness} of the corresponding prototype in the model learning.
Eventually, we impose all prototypes in $\mathcal{P}$ by the calibration factors as:
\vspace{-2mm}
\begin{equation}
    \hat{\mathcal{P}} =\{
     \text{log} (\bar\omega _c) + p_c \ | \ p_c \in \mathcal{P} 
    \}.
\end{equation}

\subsection{Proto-loss for Training}
\label{sec:loss}
To integrate the contrastive learning into our work in an end-to-end manner, we refer to the concept of  InfoNCE\cite{oord2018representation}. Unfortunately, it only supports single positive pair, which is incompatible to our work. 
Therefore, inspired by the unified contrastive loss~\cite{dai2021unimoco}, we extend it to include prototypes and involve both positive and negative adversarial proto-instance for training, which ensures the  optimization consistency of supervised and contrastive training so as to achieve decent performance compared with the former studies.

Formally, considering $\mathcal{H}^- = \mathcal{Q}^- \cup \hat{\mathcal{Q}}^-_\gamma$ as negative set, and $\mathcal{H}^+ = \mathcal{Q}^+ \cup \hat{\mathcal{Q}}^+_\gamma$ as positive set. For an instance $x_j$, the proto-loss is formulated as:

\begin{equation}
\small
\begin{aligned}
\mathcal{L}_{j}
= \text{log} \left[ 1+ \left(\sum_{ s_i^- \in \mathcal{H}^-} e^{\mathcal{G}(x_j)^\top s_i^-} + \sum_{p_i^- \in \hat{\mathcal{P}}^-} e^{\mathcal{F}(x_j)^\top p_i^-}  \right)\right. \\
\phantom{=\;\;} \quad\quad\left . \cdot \left(\sum_{ s_i^+ \in \mathcal{H}^+ } e^{-\mathcal{G}(x_j)^\top s_i^+} + \sum_{p_i^+ \in \hat{\mathcal{P}}^+} e^{-\mathcal{F}(x_j)^\top p_i^+}  \right)  \right]
.
\end{aligned}
\label{UniCon}
\end{equation}
Note that the proposed proto-loss can also be applied to general contrastive settings.




\section{Experiments}

\subsection{Datasets and Evaluation}

We conduct the experiment on two publicly available datasets. The ISIC2018 is accessed by the Skin Lesion Analysis Toward Melanoma Detection 2018 challenge~\cite{codella2018skin}. The other dataset is APTOS2019, which is provided by ~\cite{aptos2019}. For a better illustration, the details of datasets are listed in Table~\ref{tab1}. Notably, the imbalance ratio denotes as $N_{max}/N_{min}$, where $N$ is the number of samples in each class. We follow the same protocol in \cite{marrakchi2021fighting} and randomly split the original dataset into train and test sets with the ratio of 7:3. All experimental results will be reported with the criterion of accuracy and F1-score.

\vspace{-0.5cm}
\begin{table}[]
 \begin{center}
 \caption{The details of long-tailed medical datasets.}\label{tab1}
\begin{tabular}{c|c|c|c}
\toprule[1pt]
Dataset   & \# of classes & \# of samples & Imbalance ratio \\
\toprule
ISIC2018  & 7             & 10015         & 58              \\
APTOS2019 & 5             & 3662          & 10              \\
\toprule[1pt]
\end{tabular}
\end{center}
\end{table}

\vspace{-1cm}
In this work, we selected 9 existing methods as the comparison methods. 
To be specific, we use cross-entropy (CE) as the baseline.
Further, based on CE, the balanced resampling strategy is used to address the imbalanced classification problem. 
The contrastive learning method can be treated as another comparison method.
By integrating the balanced resampling strategy,``CL+resample" is proposed in \cite{marrakchi2021fighting}, which is a two-stage approach by training the backbone firstly, then freezing the backbone and training the classifier with balanced sampling. 
To down-weight easy negatives in one-stage detector, focal loss \cite{lin2017focal} is also useful in classification problem.
LDAM \cite{cao2019learning} focuses on label distribution margin, and is regarded as a simple but effective training strategy.
%
%
OHEM \cite{shrivastava2016training} is a hard negative mining method based on the model. 
DANIL \cite{gong2020distractor} explores distractors to learn better CNN features.

\begin{table}[!t]
    \centering
    \caption{Comparison with the state-of-the-art methods}\label{tab2}
\setlength{\tabcolsep}{4mm}
\setlength\extrarowheight{2pt}
    \begin{tabularx}{1.0\linewidth}{ l  >{\centering\arraybackslash}X@{}>{\centering\arraybackslash}X   >{\centering\arraybackslash}X@{}>{\centering\arraybackslash}X@{}}
   
    \toprule[1pt]
  
   \multirow{2}{*}{Methods} &  \multicolumn{2}{c}{ISIC2018} &  \multicolumn{2}{c}{APTOS2019} \\
  \cline{2-5}
                    & Accuracy             & F1-score         & Accuracy             & F1-score    \\
    \midrule
      CE          & 0.850    & 0.716             & 0.812         & 0.608      \\
    CE+resample & 0.861      & 0.735      & 0.802         & 0.583     \\
     Focal loss  & 0.849      & 0.728        & 0.815          & 0.629      \\
      LDAM        & 0.857        & 0.734       & 0.813          & 0.620    \\
  OHEM        & 0.818                & 0.660        & 0.813                 & 0.631                     \\
 MTL         & 0.811                & 0.667           & 0.813                 & 0.632          \\
 DANIL       & 0.825                & 0.674            & 0.825                 & 0.660          \\
  CL          & 0.865        & 0.739       & 0.825          & 0.652          \\
 CL+resample & 0.868        & 0.751     & 0.816          & 0.608     \\
\textbf{ProCo(ours)} & \textbf{0.887}      & \textbf{0.763}      & \textbf{0.837}      & \textbf{0.674}     \\
    \bottomrule
    \end{tabularx}
\end{table}

\subsection{Implementation Details}
The data augmentation policy and update ratio we utilized of the contrastive learning is identical to MoCoV2 \cite{chen2020improved}. ResNet50 \cite{he2016deep} is used as our backbone. We implement projector $\mathcal{G}$ with 2 fully-connected layers, of which the hidden layer size is set to 2048, followed by ReLU activation function. The projector $\mathcal{F}$ is realized by a single fully-connected layer with ReLU activation function. For simplicity, category prototype $\mathcal{P}$  can be regarded as a classifier in our framework.
The batch size is 128 and the default optimizer is SGD with a momentum of 0.9 and a weight decay of 0.0001. The initial learning rate is set to 0.05.
The initial similarity value $\omega_c$ is set to 0.01 and $\beta$ is set to 0.95. Hyperparameters $\gamma$ and $E$ are set at 20 and 0.4 via a grid search. Referring to \cite{marrakchi2021fighting}, the training epochs of ISIC2018 and APTOS2019 are set at 1,000 and 2,000, respectively. Particularly in the test phase, we only utilize $\mathcal{F}(x)$ and  $\mathcal{P}$ to acquire the prediction. 

\subsection{Comparison with the State-of-the-art}
In this part, we compare the proposed ProCo with the state-of-the-art methods on two open-release datasets: ISIC2018 and APTOS2019.
Table~\ref{tab1} presents the entire experimental results.
%
%
We can see that the proposed method achieves optimal performance regardless of the data set, demonstrating its excellent generalization.
Apart from this observation, the performance of Focal loss and LDAM is comparable to the baseline on both datasets.
In addition, OHEM, MTL, and DANIL obtain comparable accuracy and F1-score on the APTOS2019 while inferior performance on the ISIC2018, illustrating their weak generalization.
The results from the two CL-based methods are uniform and outperform the baseline.
Note that the role of the balanced resampling strategy is inconsistent across different data sets.

\vspace{-0.5cm}
 \begin{table}[!t]
 \begin{center}
 \caption{Effectiveness of each module in our ProCo framework}\label{tab3}
 \begin{tabular}{c|c|c|c|c}
 \toprule
 Proto-loss & proto-instance & prototype recalibration & Accuracy & F1-score   \\
 \hline
  $\checkmark$  &       &                    &    0.857      &   0.742       \\
 $\checkmark$       & $\checkmark$    &                      &    0.875      &   0.751      \\
  $\checkmark$       &         &    $\checkmark$       &   0.862    &     0.754       \\
  $\checkmark$    & $\checkmark$     &    $\checkmark$      &   \textbf{ 0.887 }     &  \textbf{0.763}      \\
 \toprule
 \end{tabular}
 \end{center}
 \end{table}

\subsection{Ablation Study}
In this work, the proposed framework has three fundamental modules.
To validate the effectiveness of each module, we carried out ablation studies as shown in Table~\ref{tab3}.
For this ablation study, four extra experiments have been designed by arranging and combining these three modules:
1) We discarded the proto-instance and prototype recalibration strategy and only used Proto-loss to train the proposed method.
2) Based on the above experiment setting, we introduced the proto-instance module to re-train the proposed method.
3) Integrating the prototype recalibration strategy into the first experiment setting.
4) Employing all the three modules which is our entire proposed framework.
From Table~\ref{tab3}, we can clearly observe that with all the three modules, the proposed method can obtain the best performance in terms of accuracy and F1-score as shown in the last row.
To be specific, the modified framework with only the Proto-loss as the loss function obtained an inferior performance to other three experiments.
To this end, we can benefit from proto-instance module and prototype recalibration strategy, which is consistent with the results from experiment 1 and 2. 
In addition, to evaluate the superiority of the Proto-loss over other commonly-used loss functions: cross-entropy (CE) and InfoNCE, we re-trained the proposed network using different loss functions, respectively. 
The final results of F1-score were: 0.716 (CE), 0.739 (InfoNCE), and 0.742 (Proto-loss).
We can see that using Proto-loss as the loss function will improve the final F1-score by 3.63\% compared to that from CE.



\section{Conclusion}
This paper proposes a novel paradigm called ProCo, addressing the long-tailed classification problem in a contrastive way. 
Our ProCo mainly consists of three components: i) category prototype and the adversarial proto-instance; ii) prototype recalibration strategy and iii) a unified proto-loss.
Extensive experiments on two publicly available datasets show that the efficacy of our proposed components, and our proposed framework outperforms the existing state-of-the-art long-tailed methods by a large margin.

%
%
%
\bibliographystyle{splncs04}
\bibliography{bibfile}
\end{document}